\pdfoutput=1
\documentclass[11pt]{article}

\usepackage{acl}

\usepackage{times}
\usepackage{latexsym}
\usepackage{ulem}
\usepackage{tcolorbox}
\usepackage{multirow}
\usepackage{enumitem}

\usepackage[T1]{fontenc}
\usepackage[utf8]{inputenc}

\usepackage{microtype}

\usepackage{inconsolata}

\def\dataset{\textsc{repair-qa}}

\title{\textit{No that's not what I meant}: Handling Third Position Repair in Conversational Question Answering}

\author{Vevake Balaraman~~Arash Eshghi~~Ioannis Konstas~~Ioannis Papaioannou\\
AlanaAI\\ 
\texttt{\{vevake, arash, ioannis.k, ioannis\}@alanaai.com}}

\begin{document}
\maketitle
\begin{abstract}
The ability to handle \textit{miscommunication} is crucial to robust and faithful conversational AI. People usually deal with miscommunication immediately as they detect it, using highly systematic interactional mechanisms called \textit{repair}.
One important type of repair is \textit{Third Position Repair} (TPR) whereby a speaker is initially misunderstood but then corrects the misunderstanding as it becomes apparent after the addressee's erroneous response (see Fig.\ 1).
Here, we collect and publicly release \dataset\footnote{The dataset, models and code for all experiments are available at \url{https://github.com/alanaai/Repair-QA}}, the first large dataset of TPRs in a conversational question answering (QA) setting. The data is comprised of the TPR turns, corresponding dialogue contexts, and candidate repairs of the original turn for execution of TPRs. We demonstrate the usefulness of the data by training and evaluating strong baseline models for executing TPRs.
For stand-alone TPR execution, we perform both automatic and human evaluations on a fine-tuned T5 model, as well as OpenAI's GPT-3 LLMs.
Additionally, we \textit{extrinsically} evaluate the LLMs' TPR processing capabilities in the downstream conversational QA task. The results indicate poor out-of-the-box performance on TPR's by the GPT-3 models, which then significantly improves when exposed to \dataset.
\end{abstract}

\section{Introduction}

Participants in conversation need to work together on a moment by moment basis to achieve shared understanding and coordination \cite{Clark96,Clark.Brennan91,Goodwin81,Healey.etal18,Mills07}. One of the key interactional mechanisms that enables this is called \textit{repair} \cite{Schegloff.etal77,Schegloff92} -- see Fig.\ 1: a set of universal, highly systematised \cite{Dingemanse.etal15}, local methods for dealing with \textit{miscommunication} as it is detected.

\begin{tcolorbox}[colback=yellow!10!white,colframe=blue!75!black,title=Figure 1. TPR Example from \dataset]
  \textbf{(T1)} U: What is the name of {\color{red}the princess in Frozen}? $\langle$\texttt{Trouble Source}$\rangle$\\
  \textbf{(T2)} S: The name of the princess who eventually becomes queen is Elsa\\
  \textbf{(T3)} U: {\color{blue} no I mean the name of the younger sister} $\langle$\texttt{Third Position Repair}$\rangle$\\
  \textbf{(T4)} S: The name of the younger sister is Anna

\end{tcolorbox}%\vspace{-0.5cm}

\setcounter{figure}{1} 

Miscommunication likewise arises in human-machine conversation. Therefore, the ability to interpret and generate effective repair sequences is crucial to \textit{robust} Conversational AI technology, and to ensuring that Natural Language Understanding (NLU) output and/or subsequent system responses remain \textit{faithful} to what the user intended. 

Considerable attention has been paid to computational models for the interpretation and generation of \textit{self-repair} (see \cite{Hough.Schlangen15,Hough15,Shalyminov.etal17,Skantze.Hjalmarsson10,Buss.Schlangen11,Hough.Purver12} among others): a class of repairs whereby the speaker corrects themselves on the fly within the same conversational turn (e.g.\ ``User: I want to go to London uhm sorry Paris''). Similarly, the crucial role of generating and responding to \textit{Clarification Requests} (e.g.\ ``Pardon/what/who?'') in  conversational models has long been recognised (see \cite{San-Segundo.etal01, Purver04,Purver.Ginzburg04,Rieser.Moore05,Rodriguez.Schlangen04,Rieser.Lemon06} among others), but existing systems either remain limited (e.g.\ \newcite{Curry.etal18}) or do not support this at all -- see \newcite{Purver.etal18} for an overview of existing models of repair.

In this paper, we focus on an important class of repairs that has, to our knowledge, been neglected in the NLP community, likely due to the unavailability of data: \textit{Third Position Repair} (TPR; \cite{Schegloff92}; aka repair after next turn). These occur when the addressee initially misunderstands the speaker (Fig.\ 1 at T1, the \textit{trouble source} turn), responds based on this misunderstanding (at T2), which in turn reveals the misunderstanding to the addressee who then goes on to correct the misunderstanding (at T3).
Our \textbf{contributions} are: (1) We collect, analyse and release \dataset , the first large dataset of Third Position Repairs (TPR) in a conversational QA setting together with candidate repair outcomes (rewrites) for training \textit{repair execution} models;
and (2) We then use \dataset\ to: (a) train and intrinsically evaluate strong baseline models for the execution of TPRs; and (b) systematically probe the TPR processing capabilities of GPT-3-Curie and GPT-3-Davinci with and without exposing them to examples from \dataset.

\section{The \dataset\ dataset}
In this section, we describe our method for eliciting Third Position Repairs (TPR) from AMT crowd workers (henceforth annotators). Overall, we set this up as a dialogue completion task whereby the annotators are given a dialogue snippet in which a miscommunication has occurred: they are given T1 (Fig.\ 1; the \texttt{Trouble Source}) and T2 (the erroneous system response). They are then asked to provide a (Third Position) correction at T3 to resolve the miscommunication.

\paragraph{Method: Eliciting TPRs} We built our dialogue completion tasks on Amazon Mechanical Turk (AMT). Annotators were paid $\$0.29$ per annotation for their work (estimated at $\$11$ per hour). To generate the dialogue completion tasks in order to elicit TPRs, we start from the AmbigQA dataset \cite{Min.etal20} since it contains ambiguous questions (i.e.\ questions that have multiple interpretations and answers) and their corresponding unambiguous questions along with their answers. For each ambiguous question, $Q$, and the corresponding pair of unambiguous questions with their answers, $(Q_1, A_1)$ and $(Q_2, A_2)$, we build a dialogue snippet to be completed by the annotator with a TPR as follows: (1) We build an informative \textit{context}, $C$, that differentiates between questions $Q_1$ and $Q_2$; (2) The answers in AmbigQA are mostly short, Noun Phrase answers, which do not reveal how the ambiguous question was interpreted or reveal the apparent miscommunication to the annotator. To remedy this, we transform these short answers to full sentential form using the rule-based approach of \newcite{demszky2018transforming}. This allows us to derive sentential forms for $A_1$, call it $A_1'$; (3) We build the dialogue snippet with two turns, T1 and T2 -- see Fig.\ 1 -- where $T1 = Q$ and $T2 = A_1'$. Annotators are told that their goal was to get a response to $Q_2$ (indicated by context $C$); then, given the dialogue snippet which erroneously provides an answer to $Q_1$, they are asked to provide \textit{two} alternative TPRs at $T3$ to get a response to $Q_2$ instead. For example, in Fig.\ 1: $Q$ is $T1$; $Q_1$ is ``What is the name of the princess in Frozen who eventually becomes queen?''; $A_1$ is ``Elsa''; $A_1'$ is $T2$; and $C$ is ``who eventually becomes queen vs. the younger sister''. The context $C$ is built by identifying the difference between $Q1$ and $Q2$. We employ this approach as the AmbigQA unambiguous questions have the same syntactic form as the ambiguous question. Another big advantage of using the AmbigQA dataset is that $Q_2$ can be seen as the contextually resolved meaning of the TPR which we call the gold `rewrite' following \cite{anantha-etal-2021-open}. This gold rewrite is used below in our \textit{repair execution} models.
See Appendix \ref{amt-setup} for more details.

\paragraph{Statistics and Quality Control} The \dataset\ dataset consists of \textbf{3305} examples (training: 2657, test: 648) which are chosen and annotated from the 4749 examples from the AmbigQA dataset.
Each conversation in \dataset\ consists of two different TPRs yielding a total 6610 TPR annotations. Table \ref{tab:tpr-examples} in Appendix shows some examples of the collected data. For quality control, we randomly select 100 TPR annotations from the testset to perform a qualitative inspection of the collected data. We annotate them for (i) Quality: Does the TPR convey the information needed to convey the necessary correction?; (ii) Context-Dependence: Does the TPR contain any context-dependent phenomena (e.g.\ fragments, ellipsis, pronominals); and (iii) Corrective: Is the TPR formulated explicitly as a correction? (e.g.\ The TPR in Fig.\ 1 could have been: ``what about the name of the younger sister?'' which does not explicitly signal a correction). We find that only 16\% of the data contains some noise; that 93\% of TPRs contain some form of context-dependency; and that 80\% of the TPRs formulate the TPR explicitly as a correction. To further measure the degree to which the interpretation of the TPRs relies on the dialogue context, we measure the unigram overlap between the TPR and the reference rewrite (viz.\ $Q2$ above). We find 28\% overlap between them, suggesting that the TPRs are highly context-dependent.

\paragraph{Limitations} As such, \dataset\ has two important limitations: (1) TPRs can in general sometimes -- but rarely -- occur at a distance of more than two turns from the \textit{trouble-source} turn \cite{Schegloff92}. But the TPRs we collected are always in the third turn following the trouble source: this is an artefact not just of our data collection design as a unilateral dialogue completion task, but also of the architecture of most Conversational QA models that REPAIR-QA is designed to be useful for; and (2) overall we'd have preferred a more ecologically valid setup where TPRs are elicited within a more dynamic, interactive setting rather than as a dialogue completion task. Nevertheless, we believe that this trade-off between difficulty of collecting human-human dialogues, and the breadth of the types of TPR sequences collected is justified.

\section{TPR execution}

We cast the TPR execution task as a sequence to sequence problem, where input to the model is the dialogue history up to and including the TPR turn, and the model is trained to generate a rewrite of the ambiguous, trouble-source question, reflecting the correction in the TPR. We use a pre-trained T5 model \cite{10.5555/3455716.3455856} for our experiments and compare against OpenAI's GPT-3 \cite{10.5555/3495724.3495883} when prompted with TPR examples.

\begin{table}[]
    \centering
    \begin{tabular}{c|p{9mm}|p{9mm}|p{8mm}}
        & BERT Score & BLEU & EM \\
        \hline
        \hline
        T5-\dataset & \textbf{97.48} & \textbf{72.06} & \textbf{30.40} \\
        GPT-3-Davinci & 97.22 & 64.18 & 25.68 \\
        GPT-3-Curie & 93.19 & 52.43 & 7.60 \\
    \end{tabular}    
    \caption{Model performance on the testset of the \dataset\ dataset.}
    \label{tab:rewrite-results}
\end{table}

\begin{table}[]
    \centering
    \begin{tabular}{c|c|c}
        & BERTScore & BLEU \\
        \hline
        \hline
        T5-\dataset & 1.48 & \textbf{20.12}\\
        GPT-3-Davinci & \textbf{1.76} & 19.94 \\
        GPT-3-Curie & (0.11) & 1.85 \\
    \end{tabular}    
    \caption{Model ability to generate corrective tokens computed based on the difference in performance of the prediction against the rewrite and the trouble source.}
    \label{tab:tpr-exec-diff}
\end{table}

\subsection{Repair Execution Results} 
The models are evaluated against metrics of BERTScore \cite{ZhangKWWA20}, BLEU and Exact Match (EM) between the reference rewrite and the generated output \footnote{We also tried an NLI-based text-classifier \cite{yin-etal-2019-benchmarking} for evaluation but the metric was not suited for this task, hence not reported here.}.

Table \ref{tab:rewrite-results} shows the performance of all models on the \dataset\ testset.
The T5 model is fine-tuned using the \dataset and its performance is reported as T5-\dataset.
The fine-tuned T5-\dataset\ model achieves the best performance against the gold rewrites on all the 3 metrics considered.
The GPT-3 models (Davinci and Curie) are few-shot prompted with 10 random examples, per test instance, pooled from \dataset\ followed by the test data;
(see Appendix \ref{app:prompts} for details);
unlike the T5-\dataset\ model which is \emph{fine-tuned} using the \dataset\ training data. We see a slightly lower performance for Davinci compared to the T5-\dataset\ on the automatic evaluation; the Curie model shows significantly inferior performance, especially when looking at EM \footnote{We also did a zero-shot evaluation of a T5 model trained \emph{only} on QReCC \cite{anantha-etal-2021-open} -- a contextual resolution dataset -- against the \dataset\ testset: it performed very poorly (BLEU = 37.44) indicating that the patterns of context-dependency in the TPRs are very different from the general patterns of context-dependency found in the QReCC dataset. This further demonstrates the usefulness of \dataset.}.

Generally, the correction that a TPR provides to the \textit{trouble source} question (T1 in Fig.\ 1) is very specific and small (often just 1 or 2 words, e.g. ``the younger sister'' in Fig.\ 1). Thus a higher BLEU score is more likely even when the model prediction is similar to the trouble source.
To evaluate the ability of the models to produce specifically the corrective tokens, we evaluate the models' predictions against both the gold rewrite and the trouble source itself, and compare these across all metrics. 
We compute the metrics for the models' prediction against the gold rewrite on the one hand, and, the trouble source separately on the other hand, and compute the difference between them (simple subtraction).
This difference in performance against them is therefore attributable to whether the model was able to produce the few corrective tokens. Table\ \ref{tab:tpr-exec-diff} shows this differential evaluation: a similar trend is seen on the models for the BLEU metric but GPT-3-Davinci outperforms other models on BERTScore. This result is discussed further below.

\subsection{Human Evaluation} We asked two expert annotators (two co-authors of the paper) to rate the quality of T5-\dataset\ and GPT-3-Davinci model's output rewrites for executing the TPRs. We separately asked them the following questions: \textbf{Q1}: ``On a scale of 1 to 5, how well does the model prediction avoid the misunderstanding caused by the ambiguity in the original question?''; and \textbf{Q2}: ``On a scale of 1 to 5, to what degree is the model prediction asking for the same information as the gold?''.
While the answer to Q2 depends on the gold rewrites from \dataset, the answer to Q1 does not. This is because in executing a TPR what we care about is not necessarily the surface form of the output
but instead the overall correction on a \textit{semantic level}. The annotators showed very high interannotator agreement on both questions (average Krippendorf's $\alpha = 0.8$).

\begin{table}[]\centering
    \begin{tabular}{l|l|l}
         & \textbf{Q1} & \textbf{Q2} \\
         \hline
         \hline
         T5-\dataset & 3.53 & 4.01 \\
         GPT-3-Davinci & 4.56 & 4.27 \\
    \end{tabular}
    \caption{Human evaluation of TPR execution models}
    \label{tab:human}
    % \vspace{-3mm}
\end{table}

As Table \ref{tab:human} shows, the Davinci model's performance in the human evaluation is superior to the T5-\dataset\ model for both Q1 and Q2. At first glance, this would seem to be inconsistent with the word overlap metrics in Table \ref{tab:rewrite-results} since the fine-tuned T5-\dataset\ model outputs show more overall overlap with the gold rewrites. However, a qualitative inspection of the respective outputs of each model shows that the Davinci model manages to produce rewrites which sufficiently capture the meaning of the TPR even as it doesn't always reproduce exactly the same words. This explanation is further supported by the BERTScore, semantic similarity results in Table \ref{tab:tpr-exec-diff} which shows slightly superior performance of the Davinci model
(see Table \ref{tab:model-example} in Appendix for an example comparison).
We believe that this is due to the fact the Davinci model is only exposed to ten examples in the prompt each time, whereas the T5-\dataset\ model is fine-tuned on all the training data from \dataset. 

\section{Extrinsic evaluation of GPT-3's TPR capabilities in conversational QA} \label{end-to-end eval}

\begin{table}[]
    \centering\footnotesize
    \begin{tabular}{r|c|c|c}
        Prompting & BLEU & EM & Unknown\\
         \hline
         \hline
         w/o TPR examples & 11.40 & 11.71\% & 230\\
         with TPR examples & 16.98 & 31.90\% & 57\\         
    \end{tabular}
    \caption{End-to-end, TPR processing capability of GPT-3 Davinci, with and without being exposed to TPR examples from REPAIR-QA}
    \label{tab:GPT-3-tpr-probing}
\end{table}

In this section, we use \dataset\ to evaluate the TPR processing capabilities of OpenAI's GPT-3 Davinci model extrinsically in an end-to-end, conversational QA setting. We do this by comparing: 
\begin{enumerate}[label=(\alph*)]
\item the model's response to the reference rewrite (the corrected, unambiguous form of each question); with

\item the response returned after the dialogue snippet with the TPR as its last turn. 
\end{enumerate}

If (a) and (b) are identical or highly similar, we can infer that the model was able to interpret the TPR correctly; independently of whether the responses are faithful. We compute the automatic evaluation on the model's response in (b) while treating the model's response in (a) as the ground truth. This would evaluate if the model was consistent in generating responses for both the rewrite and the TPR dialogue snippet.
This evaluation is performed under two \textit{prompting conditions}: \textbf{With TPR examples}: where the model is exposed to 10 TPR examples in the prompt; and; \textbf{Without TPR examples}: where the model is prompted without any TPR examples. In both conditions, the preamble instructs Davinci to generate \textit{unknown} as the answer if the question is either nonsense, trickery, or Davinci has no clear answer. In addition, in both cases, the model is instructed to provide short form, Noun Phrase answers (for details of all of the preambles used, see Appendix, Sec.~\ref{app:prompts}).

There could in general be two reasons for \textit{unknown} predictions after a TPR: (i) the Davinci's closed-book knowledge is insufficient to answer the (disambiguated, corrected) question; or; (ii) It was unable to interpret the TPR sequence. Since we are interested only in (ii), we \textit{exclude} all cases where the model was not able to answer the unambiguous question (i.e case (a) above), viz. the reference rewrite (the meaning of the TPR). This way we ensure that the model can actually answer the target, rewritten / corrected question. After these are excluded, the `Unknown' column in Table \ref{tab:GPT-3-tpr-probing} contains the number of \textit{unknown} responses to the TPRs;
showing how the model improves when exposed to TPR examples in conversational QA.

For cases where both (a) and (b) above receive answers from GPT3, we perform automatic evaluation to measure the similarity between them: this is also shown in Table \ref{tab:GPT-3-tpr-probing}. As a surface overlap metric, BLEU is suitable for this evaluation since we compare short answer tokens with many of these being bare Noun Phrases, e.g. names of movies, persons, dates, etc: there are no or few semantically similar paraphrases of these answers.

As is evident in Table~\ref{tab:GPT-3-tpr-probing}, the TPR processing capability of Davinci in conversational QA when not exposed to any TPR example is very poor, but this improves significantly with a handful of TPR examples in the prompt. This shows that state-of-the-art LLMs do not handle TPRs well at all out-of-the-box, validating the requirement for datasets addressing specific dialogue phenomena like TPRs. 

Even when the model is exposed to TPR sequences in the prompt (the ``with TPR examples" condition) the model's performance still leaves a lot to be desired: the model's responses to the TPRs matches the expected response only in $31.9\%$ of cases.

To verify the meaningfulness of the $31.9\%$ exact match and the corresponding low BLEU score of $16.98$ between model responses in (a) and (b), we went on to do a manual inspection of the data. Fig.~\ref{tab:e.g} shows two examples of these responses:

\begin{figure}[h]
    \small
    \centering
    \begin{tabular}{l}
        \textbf{User:} Who plays the leprechaun in the leprechaun movie? \\
        \textbf{System:} Warwick Davis \\
        \textbf{TPR:} I was referring to leprechaun origins \\
        
        \textbf{Rewrite:} Who plays the leprechaun in the Leprechaun \\
                        Origins movie? \\
        \\
        \textbf{Response to (a):} Dylan Postl \\
        \textbf{Response to (b):} Linden Porco \\
        --------------------------------------------------------------------\\
        \textbf{User:} Who created the quote keep calm and carry on? \\
        \textbf{System:} British government \\
        \textbf{TPR:} I wanted to know the name of the ministry though. \\
        \textbf{Rewrite:} Which ministry created the quote keep calm and\\
                        carry on? \\
        \\
        \textbf{Response to (a):} British Ministry of Information \\
        \textbf{Response to (b):} Ministry of Information \\ 
    \end{tabular}
    \caption{Two pairs of example responses provided by Davinci in its responses to (a): the unambiguous, corrected question rewrite; and; (b): the three turn TPR sequence}
    \label{tab:e.g}
\end{figure}

We can see different answers when prompted with the dialogue including the TPR ((b) above) and when prompted with the rewrite (unambiguous form of the input; (a) above). Such inconsistent answers are frequent from the model even when \dataset\ examples are provided in the prompt.

For more certainty, we further computed more focused BLEU scores only in cases where there was no exact match between the model's responses in (a) and (b). The BLEU scores on these not exactly matching responses, \textbf{with} and \textbf{without} exposure to TPR examples were 8.81 and 8.08 respectively. This shows that the model provides different, inconsistent answers for a large part of the \dataset\ dataset even when exposed to TPR examples in the prompt; which in turn shows that the model is not able to interpret or integrate the TPR for too large a part of \dataset. On a very small proportion of cases, Davinci provides responses which are similar (usually a partial match as in the second example above: ``British Ministry of Information'' vs. ``Ministry of Information''), which is captured by the BLEU score metric.

\section{Conclusion}
The ability to interpret and generate repairs is essential to robust and faithful Conversational AI. In this paper, we focused on Third Position Repair (TPR) that's been largely neglected in the NLP community. We collect, analyse and release the first large dataset of TPRs and use it to evaluate strong baseline repair execution models, as well as the conversational QA performance of Open AI's Davinci model when it encounters TPRs. The results show very poor out-of-the-box performance on TPRs which then improves when the model is exposed to the \dataset\ dataset. But even then, Davinci does not exhibit an acceptable performance on TPRs when evaluated end to end in a Conversational QA setting. This is a symptom of the sparsity of TPRs in the original dialogic data used to pretrain Davinci and LLMs in general; and suggests that LLM researchers should be more selective in how they compile the datasets used for pretraining.

For this paper, we did not have a chance to evaluate later releases of LLMs (e.g. GPT3.5; GPT4) - it would be telling to see how much performance improvement the later models might exhibit on TPRs. Our evaluation methods above in conjunction with the \dataset\ dataset can be used easily to perform these evaluations. Finally, we hope that this paper inspires further computational research into miscommunication phenomena in dialogue in the context of recent astonishing successes with LLMs.

\iffalse
\section{Limitations}
\begin{itemize}
    \item Limited to Question Answering; pattern of TPRs may well be different in other domains of language use (e.g.\ a multi-modal setting)
    \item The TPR execution models operate only on textual level, and therefore would not support cases where the correction needs to be made on a semantic level (e.g.\ an intent/slot representation).
    \item We do not do a user study of how different attributes of a system (e.g.\ naturalness or trustworthiness) are perceived when it is able to process TPRs.
    \item We do not provide models for recognition/classification of TPRs.
\end{itemize}
\fi
\bibliography{all}

\appendix

\begin{table*}
    \small
    \centering
    \begin{tabular}{l}
    \textbf{User:} What is the most current episode of Ray Donovan? \\
    \textbf{System:} The title of the most current episode of Ray Donovan is you'll Never Walk Alone. \\
    \textbf{User (TPR):} What number was it in the series?\\
    \\
    \textbf{GPT-3-Davinci:} What is the number of the most current episode of Ray Donovan titled\\"You'll Never Walk Alone"?\\
    \textbf{T5-QReCC+\dataset:} What number was the most current episode of Ray Donovan?\\
    \textbf{Reference:} What is the number overall of the most current episode of Ray Donovan?\\
    \end{tabular}
    \caption{Prediction from different models on an example from \dataset.}
    \label{tab:model-example}
\end{table*}

\begin{table*}
    \small
    \centering
    \begin{tabular}{l}
    \textbf{User:} Where do you hit to test your reflexes?\\
    \textbf{System:} You hit to test your ankle jerk reflexes in Achilles tenden.\\
    \\
    \textbf{TPR-1:} No, I meant your biceps, not ankle.\\
    \textbf{TPR-2:} I should have been clearer. I wanted to know about the location to test for biceps reflexes.\\
    \textbf{Rewrite:} Where do you hit to test your bicips reflexes?\\
    [1.5ex]
    \hline\\
    [-1.5ex]
    \textbf{User:} Who sings i'm telling you i'm not going?\\
    \textbf{System:} Jennifer Holliday sings i'm telling you i'm not going in the musical Dreamgirls.\\
    \\
    \textbf{TPR-1:} I should have asked, who sang the song in 1982.\\
    \textbf{TPR-2:} I wanted the singer in 1982, not in the musical Dreamgirls.\\
    \textbf{Rewrite:} Who sings i'm telling you i'm not going in 1982?\\
    [1.5ex]
    \hline\\
    [-1.5ex]
    \textbf{User:} Who is the lead singer of doobie brothers?\\
    \textbf{System:} Johnston is the first lead singer of doobie brothers.\\
    \\
    \textbf{TPR-1:} I want to know who was the second lead singer not the first.\\
    \textbf{TPR-2:} I was wanting to know the second lead singer not the first.\\
    \textbf{Rewrite:} Who is the second lead singer of doobie brothers?\\
    [1.5ex]
    \hline\\
    [-1.5ex]
    \textbf{User:} Who has won the european cup the most?\\
    \textbf{System:} Real Madrid has won the european cup the most.\\
    \\
    \textbf{TPR-1:} Instead of club, can you tell me the country with the most.\\
    \textbf{TPR-2:} I am looking for the country instead of the club with them most.\\
    \textbf{Rewrite:} What country has won the european cup the most?\\
    [1.5ex]
    \hline\\
    [-1.5ex]
    \textbf{User:} How much did titanic make at the box office?\\
    \textbf{System:} Titanic ( 1953 film) made \$2,250,000 at the box office.\\
    \\
    \textbf{TPR-1:} I meant the 1997 version.\\
    \textbf{TPR-2:} I was thinking of the 1997 one.\\
    \textbf{Rewrite:} How much did Titanic ( 1997 film) make at the box office?\\
    [1.5ex]
    \hline\\
    [-1.5ex]
    \textbf{User:} Who is winner of womens world cup 2017?\\
    \textbf{System:} New Zealand is the winner of the Women's Rugby World Cup in 2017.\\
    \\
    \textbf{TPR-1:} Yeah, but who won the cricket world cup?\\
    \textbf{TPR-2:} What I wanted to know is who won the cricket cup.\\
    \textbf{Rewrite:} Who is the winner of the Women's Cricket World Cup in 2017?\\
    [1.5ex]
    \hline\\
    [-1.5ex]
    \textbf{User:} Who plays the king of france in the borgias?\\
    \textbf{System:} Michel Muller plays King Charles VIII of France in The Borgias ( 2011 TV series).\\
    \\
    \textbf{TPR-1:} I meant to ask who played louis xii.\\
    \textbf{TPR-2:} Sorry but I was looking for louis xii.\\
    \textbf{Rewrite:} Who plays King Louis XII of France in The Borgias ( 2011 TV series)?\\
    [1.5ex]
    \hline\\
    \end{tabular}
    \caption{Examples from the \dataset\ dataset.}
    \label{tab:tpr-examples}
\end{table*}

\section{Model Training and Inference} \label{app:model-details}
The T5 models reported in this paper are implemented in pytorch using HuggingFace \cite{wolf-etal-2020-transformers} library. The hyperparameter of the models are set as default with the batch size set to 16. The T5 models are trained on a single 16GB GPU and fine-tuned for 5 epochs. The results in Table \ref{tab:rewrite-results} for T5 models for a single run on the train/test split.
For GPT-3 inference, we use OpenAI's playground \footnote{https://beta.openai.com/playground} API and get predictions from both Davinci (text-davinci-003) and Curie (text-curie-001) models.

\section{Data Collection Details}\label{amt-setup}
We use Amazon Mechanical Turk\footnote{www.mturk.com} for collecting the human annotations for TPR. The data collection was conducted anonymously.

\paragraph{Crowdworker Quality Control.} We conduct a pilot with 4 internal annotators to verify the instructions and revise them before deploying to AMT crowdworkers. To control for the quality of annotations and the langauge, the crowdworkers are restricted to i) Location is one of Australia, Canada, New Zealand, United Kingdom, United States; ii) HIT approval rate > 80\% and; ii) Number of HITs approved > 50. This was done explicitly to control the quality of the annotations collected after examining the annotations from a pilot phase in AMT.

\begin{figure*}
  \includegraphics[width=\textwidth]{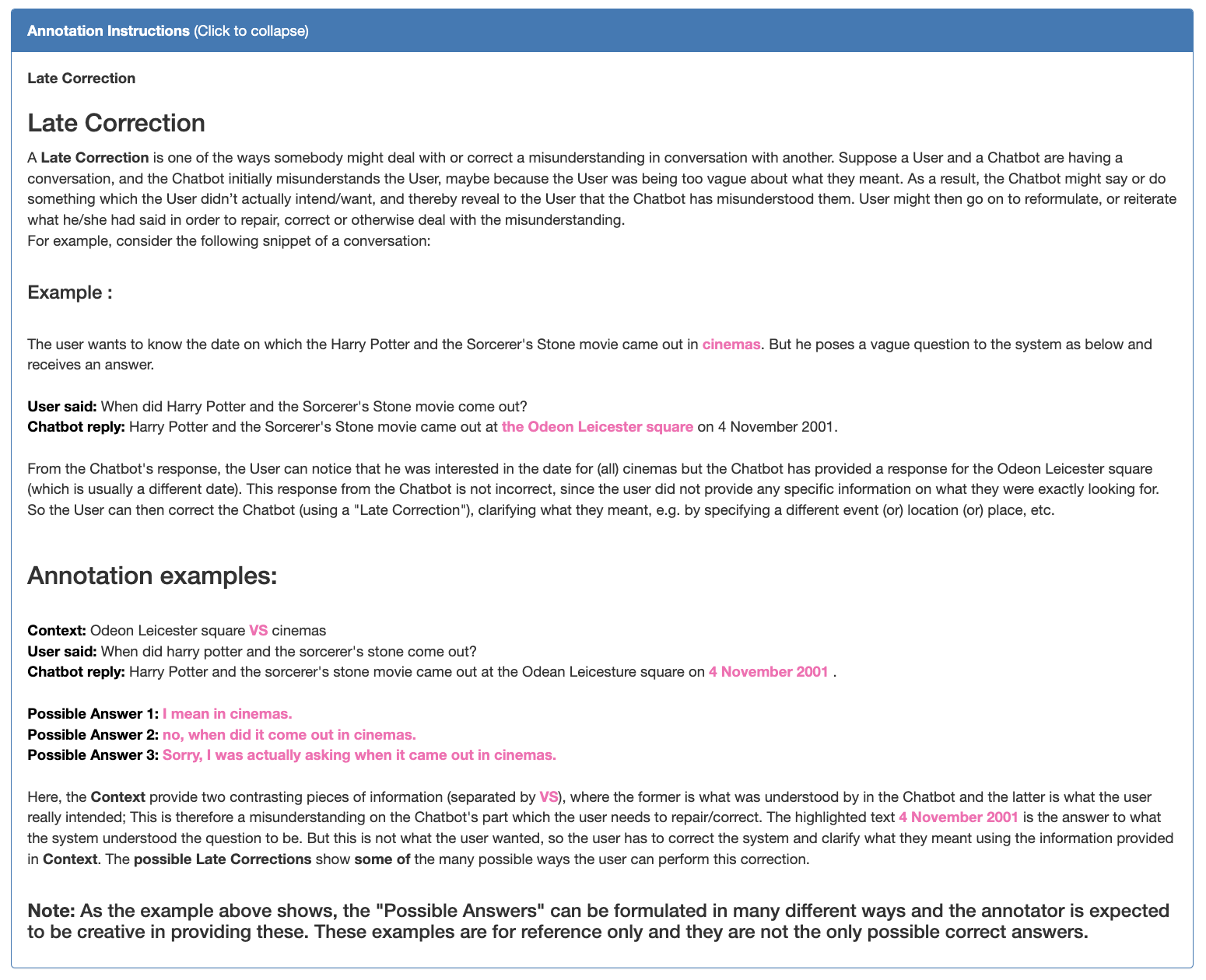}
  \caption{Annotation Instructions provided to the crowd annotators.}
  \label{fig:amt-instruction}
\end{figure*}

\begin{figure*}
  \includegraphics[width=\textwidth]{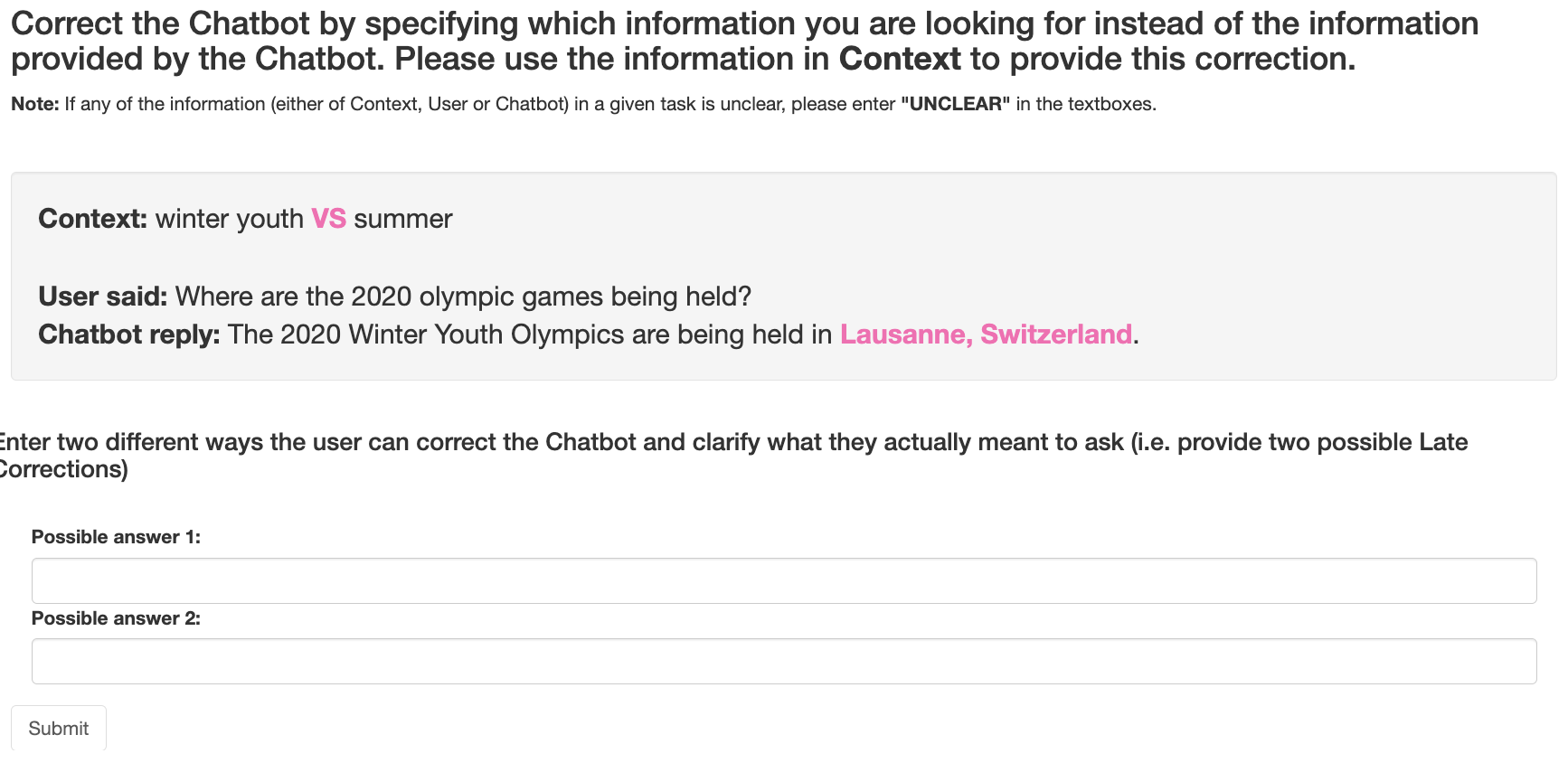}
  \caption{Interface of the annotation page as shown to the annotators.}
  \label{fig:amt-hit}
\end{figure*}

\paragraph{Crowdworker Instructions.} Figure \ref{fig:amt-instruction} shows the instruction provided to the crowdworker and Figure \ref{fig:amt-hit} shows the interface, which the crowdworker uses to annotate the provided example. We explicitly instruct the crowdworkers to mark examples in which any of the information is unclear. To better explain the concept of TPR to the crowdworkers, we use the term \textbf{late correction instead of TPR} in the annotation instructions.


\section{GPT-3 prompts} \label{app:prompts}
The prompt used to query GPT-3 model to get predictions for both rewrite and QA is presented here. The {\color{blue}{text in blue}} indicate the tokens that the GPT-3 has to generate.

\paragraph{Rewriter prompts.} Prompt used to generate rewrites from GPT-3. We use 5 examples in the prompts (single example is shown here for reference).\\
"Rewrite the Question Q based on the late correction LC.\\
\\
Q: What is the percentage of agriculture in gdp of india?\\
A: The percentage of agriculture in gdp of india in 2017 is 15.4.\\
LC: I am looking for the year 2014 instead.\\
Rewrite: What is the percentage of agriculture in gdp of india in 2014?\\
\\
Q: Who sang the song it's the final countdown?\\
A: Europe was the band that sang the song it's the final countdown, released in 1986.\\
LC: I was looking for the name of the lead singer.\\
Rewrite: {\color{blue}{Who sang lead vocals for the song it's the final countdown, released in 1986?}}\\
"

\paragraph{QA prompts.} The prompt used for the conversational QA task is as below. We use 10 examples in the prompts (single example is shown here for reference).\\
"I am a highly intelligent question answering bot. If you ask me a question that is rooted in truth, I will give you only the answer phrase. If you ask me a question that is nonsense, trickery, or has no clear answer, I will respond with "Unknown".\\
\\
Q: Who is the lead singer of doobie brothers?\\
A: Johnston is the first lead singer of doobie brothers.\\
Q: I want to know who was the second lead singer not the first.\\
A: Michael McDonald\\
\\
Q: Who sang dedicated to the one i love?\\
A: The Shirelles sang Dedicated to the one I love in 1959.\\
Q: Could you also tell me who sang the 1967 version of dedicated to the one I love?\\
A: {\color{blue}{The Mamas and the Papas}}\\
"

\end{document}